\documentclass{ceurart}

\usepackage{array}
\usepackage{booktabs}
\usepackage{listings}
\usepackage{algorithm}      
\usepackage{algpseudocode}  
\usepackage{makecell}
\usepackage{xcolor}

\usepackage[most]{tcolorbox}

\newtcolorbox{promptbox}[1]{
  enhanced,
  breakable,
  colback=white,
  colframe=black!80,
  colbacktitle=black!80,
  coltitle=white,
  title={#1},
  fonttitle=\bfseries\large,
  fontupper=\small,
  boxrule=0.8pt,
  arc=2mm,
  outer arc=2mm,
  left=4mm,
  right=4mm,
  top=3mm,
  bottom=3mm,
  before skip=8pt,
  after skip=8pt
}

\providecommand{\traceph}[1]{%
  \texttt{\textless#1\textgreater}%
}

\usepackage{newunicodechar}
\newunicodechar{→}{\ensuremath{\rightarrow}}
\newunicodechar{←}{\ensuremath{\leftarrow}}
\newunicodechar{↓}{\ensuremath{\downarrow}}
\newunicodechar{↑}{\ensuremath{\uparrow}}
\newunicodechar{≈}{\ensuremath{\approx}}
\newunicodechar{×}{\ensuremath{\times}}
\newunicodechar{÷}{\ensuremath{\div}}
\newunicodechar{≥}{\ensuremath{\geq}}
\newunicodechar{≤}{\ensuremath{\leq}}
\newunicodechar{±}{\ensuremath{\pm}}
\newunicodechar{−}{\ensuremath{-}}   
\newunicodechar{§}{\S}
\newunicodechar{•}{\textbullet}
\newunicodechar{°}{\ensuremath{^\circ}}
\newunicodechar{…}{\ldots}

\graphicspath{{figures/}}

\begin{document}

\copyrightyear{2026}
\copyrightclause{Copyright for this paper by its authors.
  Use permitted under Creative Commons License Attribution 4.0
  International (CC BY 4.0).}

\conference{}

\title{TRACE: Agentic Catalog Enrichment with Multi-source Evidence Grounding}

\author[1]{Rohan Kumar}[email=rohan.kumar@doordash.com]
\author[1]{Steven Xu}[email=steven.xu@doordash.com]
\author[1]{Kyle MacDonald}[email=kyle.macdonald@doordash.com]
\author[1]{Matthew Long}[email=matthew.long@doordash.com]
\author[1]{Bernice Chow}[email=bernice.chow@doordash.com]
\author[1]{Mac VanRenterghem}[email=mac.vanrenterghem@doordash.com]
\author[1]{Sudeep Das}[email=sudeep.das2@doordash.com]
\address[1]{DoorDash, Inc.}

\begin{abstract}
Product catalogs underpin search, discovery, and recommendation in e-commerce,
yet they are often attribute-sparse: the attributes shoppers and downstream
systems rely on are either buried in unstructured content such as titles and
images or missing from the catalog altogether. Manually enriching e-commerce
catalogs is impractical given their scale and rapid growth.
This paper introduces TRACE, a novel framework for automated catalog attribute
enrichment using agentic Large Language Models (LLMs). A \texttt{ScoutAgent} triangulates multimodal evidence across merchant catalogs, syndicated feeds, and identity-matched web search to propose candidate attribute values with supporting evidence, while a \texttt{JudgeAgent} verifies the proposed value for each attribute value against its supporting evidence and decides whether to publish it or route it to human review. On an offline human evaluation dataset, TRACE's proposed attribute values were 98.2\% accurate at 74.7\% attribute coverage. Deployed in production on an industry-scale catalog, TRACE increased impression-weighted enrichment coverage across four business verticals by 90.4\%. An online experiment subsequently showed that surfacing the enriched attributes on the product detail page increased checkout conversion by 0.48\%.
\end{abstract}

\begin{keywords}
  catalog enrichment \sep
  multimodal LLM \sep
  attribute extraction \sep
  LLM-as-judge \sep
  agentic search
\end{keywords}


\maketitle

\section{Introduction}
\label{sec:intro}

E-commerce and marketplace platforms rely on product catalogs that combine unstructured content (e.g., titles and images) with structured attributes (e.g., brand, size, and dietary tags). Structured attributes help shoppers filter and compare products, enable search systems to match queries to attribute intents~\cite{zhang2021queaco, nigam2019semantic}, and allow recommendation models to learn granular product affinities~\cite{wang2019kgat}. In practice, catalog data provided by sellers and merchants is often attribute-sparse~\cite{yang2022mave, dong2020autoknow}. Relevant values may be buried in unstructured content rather than represented as structured fields, or missing from the catalog altogether. As a result, shoppers may lack the information needed to make confident purchasing decisions or form accurate expectations about the products they receive, while downstream systems must rely on coarser product representations, limiting retrieval precision and the depth of personalization. Catalog enrichment seeks to close these gaps by extracting or sourcing missing attribute values and converting them into structured fields. Performing this work manually requires parsing free text, transcribing images, and triangulating evidence across data sources with heterogeneous formats and varying quality, making it error-prone and infeasible for large-scale, rapidly growing product catalogs.

Early work on product attribute extraction has relied on task-specific natural language processing (NLP) models trained to recover structured values from product text and images~\cite{zheng2018opentag, zhu2020mjave}. More recently, large language models (LLMs) have enabled more flexible approaches through their zero-shot and few-shot capabilities, reducing the need for labeled data and task-specific fine-tuning~\cite{brinkmann2023llmpave, sinha2024pae}. Despite this advancement, two practical challenges remain: (1) Some attribute values cannot be reliably inferred from owned data sources and must instead be sourced externally and verified against the exact product. (2) Accuracy estimated from a point-in-time catalog audit may become less representative as the catalog's product mix evolves. As a result, a system that performs well on average may still publish unsupported or product-mismatched values, particularly for products and evidence patterns underrepresented in the evaluation sample. Missing or inaccurate attribute values can mislead shoppers and degrade fulfillment quality; for safety-sensitive attributes such as allergens or dietary restrictions, they can have especially serious consequences.

\begin{figure}[t]
  \centering
  \includegraphics[
    width=\linewidth,
    trim={0 90pt 0 120pt},
    clip
  ]{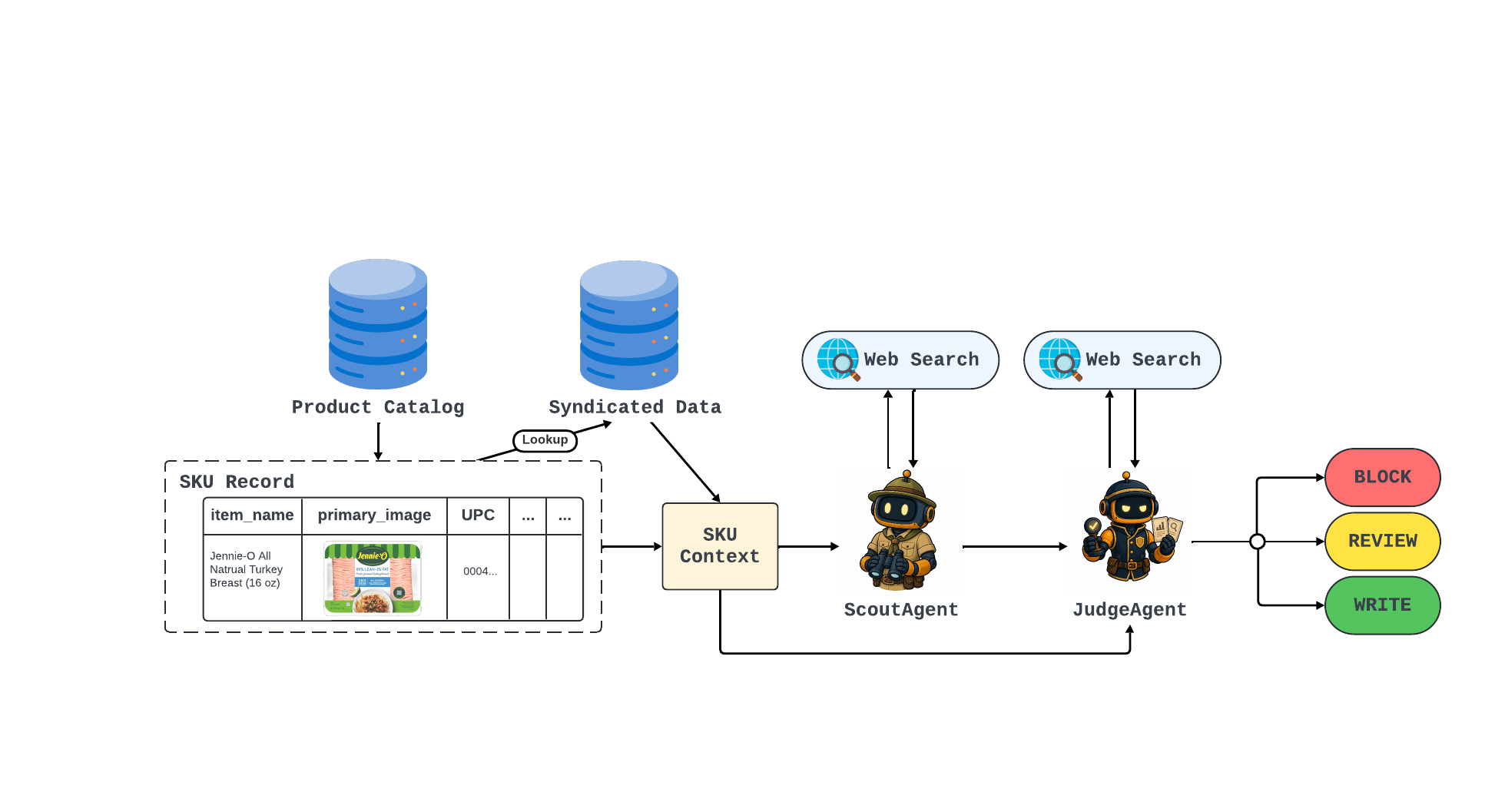}
  \caption{TRACE's architecture. The
  \texttt{ScoutAgent} receives the SKU record together with available
  catalog, syndicated, and image evidence. It may perform iterative,
  identity-matched web searches to resolve missing information and
  produces candidate attribute values with supporting evidence and
  provenance. The \texttt{JudgeAgent} independently verifies each
  candidate, optionally conducting additional web searches, and
  assigns a per-attribute verdict. The write gate then routes each
  proposed value to \texttt{WRITE}, \texttt{BLOCK}, or
  \texttt{REVIEW}.}
  \label{fig:arch}
\end{figure}

To address these challenges, we present TRACE (Figure~\ref{fig:arch}), a novel framework for automated catalog attribute enrichment using agentic Large Language Models (LLMs). The workflow is divided among two specialized agents: a \texttt{ScoutAgent} that gathers grounded evidence from multiple sources, and a \texttt{JudgeAgent} that verifies proposed value before publication. Our main contributions are threefold: (1) We introduce an end-to-end agentic LLM framework for industry-scale product catalog enrichment. The workflow triangulates heterogeneous evidence across multiple data sources while preserving provenance. (2) We incorporate identity-matched search grounding to recover attribute values that cannot be reliably inferred from owned data sources. (3) We place a \texttt{JudgeAgent} in the serving path as a verify-before-write gate, applying a consistent evidence standard to each proposed value and making publication quality less sensitive to shifts in the catalog's product mix. Through offline evaluation on datasets spanning four marketplace verticals, we show that TRACE can enrich attribute-sparse product catalogs with expert-level accuracy while achieving a scale and throughput unattainable through manual enrichment. 

\section{Related Work}
\label{sec:related}

Recovering structured attributes from unstructured text and images using
generative models is a well-studied problem. Early approaches fine-tuned
encoder-decoder models for attribute generation~\cite{khandelwal2023generative}.
More recent methods use foundation models through zero-shot or few-shot
prompting~\cite{sinha2024pae}, sometimes augmented with retrieval over similar
catalog entries~\cite{zhang2025patternrag}. Multi-agent systems are also
emerging, with recent work using iterative collaboration to refine attribute
predictions and cross-check extracted knowledge~\cite{huang2025madiave,lu2025karma}.
These methods generally assume that the target value can be inferred from owned
data sources, and the attribute of interest has
already been extracted for similar products.

LLM-as-judge methods are widely used to evaluate model
outputs~\cite{zheng2023mtbench,liu2023geval}, including in grounded
settings~\cite{saadfalcon2024ares,es2024ragas} and multi-model juries designed
to reduce single-model bias~\cite{verga2024poll}. They may serve as standalone
evaluators or as critics within multi-agent workflows. However, little prior work has examined how an LLM judge can be integrated into a production catalog-enrichment pipeline to make per-value publication decisions under varying evidence quality and a continually changing product and attribute mix.

External search grounding helps LLMs access specialized or up-to-date knowledge
beyond the provided data. Prior work retrieves real-time
information to ground generation~\cite{shi2025searchrag} and uses agentic,
tool-augmented search, in which a model interleaves reasoning with search
actions to gather evidence and reduce hallucination~\cite{yao2023react}. These methods are not directly suited to catalog enrichment, where topical relevance alone is insufficient: a retrieved source may describe a closely related product or variant with different attribute values. Sourcing catalog attributes therefore requires explicit identity matching to ensure that the evidence applies to the exact product being enriched.

\section{Methodology}
\label{sec:arch}

\subsection{Overview}
\label{sec:method_overview}

Let \(s\) denote a stock-keeping unit (SKU), \(x_s\) its catalog
record, and \(c_s\) its leaf category. A category-specific template
maps \(c_s\) to a set of priority attributes:
\begin{equation}
    \mathcal{A}_s = \mathrm{Template}(c_s; \mathcal{T}),
\end{equation}
where \(\mathcal{T}\) is the collection of attribute templates.
These templates restrict enrichment to attributes that are meaningful
for the product category.

For each attribute \(a \in \mathcal{A}_s\), the objective is to
produce either a grounded candidate value or an explicit abstention~\cite{wen2025knowlimitssurveyabstention}.
We represent a candidate as
\begin{equation}
    c_a =
    \left(
        a,\,
        v_a,\,
        E_a,\,
        \tau_a,\,
        q_a,\,
        z_a
    \right),
\end{equation}
where \(v_a\) is the proposed value, \(E_a\) is its supporting
evidence, \(\tau_a\) records the evidence-source types, \(q_a\) is
a model-reported confidence score, and \(z_a\)
records the extraction status as one of \texttt{extracted},
\texttt{not\_found}, \texttt{not\_applicable},
\texttt{ambiguous}, or \texttt{conflict}. The final system output
maps each candidate to one of three operational actions:
\texttt{WRITE}, \texttt{BLOCK}, or \texttt{REVIEW}.

\textsc{TRACE} implements this process as a two-stage
verify-before-write architecture. The \texttt{ScoutAgent} gathers
evidence from multiple sources, verifies that externally retrieved
evidence refers to the target product, and proposes grounded
attribute values. The \texttt{JudgeAgent} then re-examines each
candidate under a stricter verification policy and determines whether
it is eligible for publication, should be blocked, or requires human
review. Figure~\ref{fig:arch} shows the end-to-end
workflow, and Algorithm~\ref{alg:trace} formalizes the procedure.

\begin{algorithm}[t]
\caption{TRACE enrichment for a SKU $s$.}
\label{alg:trace}
\begin{algorithmic}[1]
\Require SKU $s$, category templates $\mathcal{T}$, threshold $\theta$
\Ensure Candidate values with actions in
$\{\textsc{write},\textsc{block},\textsc{review}\}$

\State $\mathcal{A}_s \gets \mathrm{Template}(s,\mathcal{T})$
\If{$\mathcal{A}_s = \emptyset$}
    \State \Return $\emptyset$
\EndIf

\State $E \gets
\mathrm{Catalog}(s)
\cup \mathrm{Syndicated}(s)
\cup \mathrm{Image}(s)$

\ForAll{$a \in \mathcal{A}_s$ unresolved by $E$}
    \ForAll{$p \in \mathrm{Search}(\mathrm{Query}(s,a))$}
        \If{$\mathrm{IdentityMatch}(p,s)$}
            \State $E \gets E \cup \mathrm{ExtractEvidence}(p,a)$
        \EndIf
    \EndFor
\EndFor

\State $C \gets \mathrm{Reconcile}(\mathcal{A}_s,E)$
\Comment{normalize, merge, or abstain}

\State $Y \gets \mathrm{\texttt{JudgeAgent}}(C,E)$
\State $d \gets \mathrm{WritePolicy}(Y,C,\mathrm{confidence},\theta)$
\State \Call{Apply}{$d,c$}

\State \Return $C$
\end{algorithmic}
\end{algorithm}

\subsection{The ScoutAgent}
\label{sec:extraction}

The \texttt{ScoutAgent} gathers and reconciles evidence for each target
attribute in two stages. It first considers readily available,
product-linked information, including textual fields and product
images from seller-provided catalog data and syndicated product
data. When this information is insufficient to determine a reliable
value for the target attribute, the \texttt{ScoutAgent} uses web search to
gather additional evidence.

\begin{enumerate}
    \item \textbf{Seller-provided catalog data.}
    Structured and unstructured product information supplied by the
    seller, including textual fields (e.g., item name, description)
    and product images.

    \item \textbf{Syndicated product data.}
    Product records supplied by commercial data syndicators. These
    records can provide authoritative specifications and product
    images, although their coverage and mapping quality vary.

    \item \textbf{Agentic web search.}
    For attributes that remain unresolved, the \texttt{ScoutAgent} constructs
    targeted queries from the available product identifiers and the
    target attribute to retrieve additional evidence from the web.
\end{enumerate}

Although product images originate from seller-provided or syndicated
records, TRACE treats image evidence as a separate tier from textual
evidence because visual attribute extraction is generally noisier
than extraction from text. This separation allows the \texttt{ScoutAgent} to
prioritize textual evidence while still using images for attributes
expressed visually, such as material, certification marks, and
on-package claims.

\paragraph{Identity-grounded web retrieval.} Web search may return pages that appear relevant to the query but
refer to a different product or product variant. Query relevance
alone is therefore insufficient for catalog enrichment, where the
evidence must apply to the exact product being enriched. Before
extracting an attribute value, the \texttt{ScoutAgent} verifies that the
retrieved page refers to the target product. In our implementation, identity matching is performed within the
\texttt{ScoutAgent}'s ReAct-style reasoning loop~\cite{yao2023react} using
the available product signals, including identifiers and descriptive
metadata in the catalog record and retrieved page. In
Algorithm~\ref{alg:trace}, \(\mathrm{IdentityMatch}(p,s)\) denotes
this internal reasoning step for a retrieved page \(p\) and SKU
\(s\), rather than a separate model call. Evidence from \(p\) is used
only when the \texttt{ScoutAgent} determines that the page describes the
target product; otherwise, the page is discarded.

\paragraph{Evidence reconciliation and abstention.}
After gathering evidence, the \texttt{ScoutAgent} normalizes and reconciles
candidate values across sources. It maps benign variations to a
common representation, such as ``NiMH'' and
``nickel-metal hydride,'' or ``60 Hz'' and ``60Hz.'' Evidence
supporting the same normalized value is then consolidated into a
single candidate while preserving source-level provenance. When the available evidence is insufficient, ambiguous, or
conflicting, the \texttt{ScoutAgent} abstains rather than inferring a value
from background knowledge. It records the outcome as one of
\texttt{extracted}, \texttt{not\_found},
\texttt{not\_applicable}, \texttt{ambiguous}, or
\texttt{conflict}. For each target attribute, the \texttt{ScoutAgent} outputs a candidate record
containing the normalized value when available, supporting evidence,
source types, model-reported confidence, and extraction status. The
candidate record and its provenance are then passed to the \texttt{JudgeAgent}
for verification.

The prompt template for \texttt{ScoutAgent} is provided in Figure~\ref{fig:scout-prompt} in Appendix.

\subsection{The JudgeAgent}
\label{sec:judge-gate}

Before a candidate can be written to the catalog, it is adjudicated
by the \texttt{JudgeAgent}. For each target attribute, the \texttt{JudgeAgent} receives
the candidate record produced by the \texttt{ScoutAgent}, including the
proposed value, supporting evidence, source types, and extraction
status. It reassesses whether the evidence applies to the target
product and whether it supports the proposed value. Similar to the \texttt{ScoutAgent}, the \texttt{JudgeAgent}
may use web search to gather additional evidence when it determines the supplied evidence is insufficient for verification.

The \texttt{JudgeAgent} operates at the SKU level, allowing candidates
to be evaluated in parallel across SKUs. It applies a
stricter evidence policy than \texttt{ScoutAgent}, focusing on whether 
available evidence supports the proposed value for the target
product.

\paragraph{Verdict taxonomy.}
For each candidate, the \texttt{JudgeAgent} returns one of four verdicts:

\begin{itemize}
    \item \texttt{PASS}: The available evidence supports the proposed
    value for the target product.

    \item \texttt{FAIL}: The proposed value is contradicted by the
    evidence or is not supported by it. The verdict includes a
    diagnostic subtype, such as \texttt{HALLUCINATION} or \texttt{CONTRADICTION}.

    \item \texttt{UNVERIFIED}: The proposed value is not contradicted,
    but the available evidence does not directly confirm it for the
    target product. This verdict represents insufficient verification
    rather than an identified error.

    \item \texttt{UNCERTAIN}: The evidence is conflicting or
    ambiguous, preventing the \texttt{JudgeAgent} from reaching a reliable
    determination.
\end{itemize}

The distinction between \texttt{UNVERIFIED} and \texttt{UNCERTAIN}
separates a lack of confirming evidence from active disagreement
among the available evidence. The empirical motivation for this
distinction is discussed in Section~\ref{sec:results}.

\paragraph{From verdict to catalog action.}
The \texttt{JudgeAgent} verdict is separated from the operational write
policy. Candidates below the
model-reported confidence threshold \(\theta\) are blocked. Among
the remaining candidates, those receiving \texttt{PASS} or
\texttt{UNVERIFIED} are written, those receiving \texttt{FAIL} are
blocked, and those receiving \texttt{UNCERTAIN} are routed to human
review together with their evidence trail. Formally,
\begin{equation}
d_a =
\begin{cases}
    \texttt{BLOCK},  & q_a < \theta
        \text{ or } y_a = \texttt{FAIL}, \\
    \texttt{REVIEW}, & y_a = \texttt{UNCERTAIN}, \\
    \texttt{WRITE},  & y_a \in
        \{\texttt{PASS},\texttt{UNVERIFIED}\},
\end{cases}
\end{equation}
where \(y_a\) is the \texttt{JudgeAgent} verdict and \(q_a\) is the
\texttt{ScoutAgent}'s model-reported confidence. Separating the verdict from the write policy allows
the publication rules to be adjusted without changing the
\texttt{JudgeAgent}'s verdict taxonomy.

The prompt template for \texttt{JudgeAgent} is provided in Figure~\ref{fig:judge-prompt} in Appendix.

\section{Experiments}
\label{sec:exp}

We conduct comprehensive experiments to evaluate TRACE, our agentic framework for catalog enrichment, using data sampled from the production catalog of an
e-commerce marketplace platform spanning multiple business
verticals. We combine human evaluation with \texttt{JudgeAgent}-based
adjudication to assess the quality of the enriched catalog. 

Unless otherwise stated, all experimental results use Gemini 2.5~\cite{comanici2025gemini25pushingfrontier}
Flash as the backbone for both the \texttt{ScoutAgent} and \texttt{JudgeAgent}. We compare alternative VLM backbones for \texttt{ScoutAgent} in Section~\ref{sec:ablation}.

\subsection{Data Collection}
\label{sec:data}
We evaluate TRACE on products from four business verticals:
Grocery, Alcohol, Electronics, and Home Improvement. The number of
distinct target attributes ranges from 11 in Grocery to 409 in Home
Improvement.

\paragraph{Grocery and Alcohol.}
This dataset contains 500 SKUs and 2,497 target SKU--attribute
pairs. Human annotators established the reference attribute values,
and a separate group of auditors reviewed the values proposed by
the \texttt{ScoutAgent}. We use this dataset to measure human-validated
extraction quality and to analyze the behavior of the \texttt{JudgeAgent}
against human judgments.

\paragraph{Electronics and Home Improvement.}
This dataset contains 955 SKUs and 4,990 target SKU--attribute
pairs. Because exhaustive human labeling was not available for
these verticals, we use the \texttt{JudgeAgent} to adjudicate the complete
dataset. The \texttt{JudgeAgent} results provide a scalable
operational quality signal.

\subsection{Evaluation Metrics}
\label{sec:metrics}
We evaluate the \texttt{ScoutAgent} using extraction accuracy and attribute
coverage. Extraction accuracy is the fraction of extracted values
judged correct. We report \emph{human-validated accuracy} when
correctness is determined by human review. Attribute coverage is
the fraction of requested SKU--attribute pairs for which the
\texttt{ScoutAgent} produces a nonempty value.

For datasets evaluated with the \texttt{JudgeAgent}, we report the
distribution of \texttt{PASS}, \texttt{UNVERIFIED}, \texttt{FAIL},
and \texttt{UNCERTAIN} verdicts. In particular, we refer to the
fraction receiving \texttt{PASS} or \texttt{UNVERIFIED} as the
\emph{judge-supported rate}. This metric measures compliance with
the \texttt{JudgeAgent}'s evidence policy and is not interpreted as
human-validated accuracy.

\subsection{Evaluation Results}
\label{sec:results}
\paragraph{Grocery and Alcohol.}
On the fully human-labeled Grocery and Alcohol dataset, the
\texttt{ScoutAgent} achieved 98.2\% extraction accuracy at 74.7\% attribute
coverage.

We additionally use the human labels to analyze \texttt{JudgeAgent} behavior
on the values produced by the \texttt{ScoutAgent}. Unpopulated attributes are
excluded because only proposed values enter the production
verification and write gate. During early development, the
\texttt{JudgeAgent} assigned each proposed value either \texttt{PASS} or
\texttt{FAIL}. This binary formulation grouped together several
distinct reasons for withholding approval, including explicit
contradiction, insufficient evidence for verification, and
conflicting or ambiguous evidence.

Among values assigned \texttt{PASS}, 98.4\% were confirmed correct
by human reviewers. Of the disagreements between the \texttt{JudgeAgent} and
human reviewers, 87.8\% were false rejections --- values assigned
\texttt{FAIL} but judged correct by humans --- whereas 12.2\% were
false acceptances. The binary policy therefore achieved high
precision among approved values but lower recall on correct values,
reflecting an overly conservative rejection policy.

This asymmetry motivated the current four-verdict taxonomy. The
evidence requirement for \texttt{PASS} remains unchanged, while the
former rejection outcome is divided into \texttt{FAIL} for
contradicted or unsupported values, \texttt{UNVERIFIED} for
plausible values that cannot be directly confirmed, and
\texttt{UNCERTAIN} for cases with conflicting or ambiguous evidence.
Given the high precision observed among approved values, we use the
current \texttt{JudgeAgent} as a scalable operational audit signal and report
judge-based results separately from human-validated accuracy.

\paragraph{Electronics and Home Improvement.}
On the Electronics and Home Improvement dataset, the \texttt{ScoutAgent}
achieved 87.8\% attribute coverage. Of the extracted values, 97.4\%
received a \texttt{PASS} or \texttt{UNVERIFIED} verdict from the
\texttt{JudgeAgent}.

\subsection{LLM Backbone Comparison}
\label{sec:backbone}
\label{sec:ablation}

We compare three alternative VLM backbones for the \texttt{ScoutAgent}
against the Gemini 2.5 Flash baseline, while holding the \texttt{JudgeAgent}
and all other pipeline components fixed.

\begin{table}[t]
  \caption{\texttt{ScoutAgent} backbone comparison on the Electronics and
  Home Improvement dataset, with the \texttt{JudgeAgent} fixed to Gemini 2.5
  Flash. Judge-supported rate is the fraction of all extracted values
  receiving a \texttt{PASS} or \texttt{UNVERIFIED} verdict;
  \texttt{UNCERTAIN} and invalid responses remain in the denominator.
  Publication coverage is the fraction of requested attributes
  receiving one of these two verdicts. Costs are normalized to
  Gemini 2.5 Flash.}
  \label{tab:backbone}
  \centering
  \footnotesize
  \setlength{\tabcolsep}{2.5pt}
  \begin{tabular}{@{}lcccc@{}}
    \toprule
    VLM backbone
      & \makecell{Judge-supported rate}
      & \makecell{Attribute coverage}
      & \makecell{Publication coverage}
      & \makecell{Relative cost} \\
    \midrule
    Gemini 2.5 Flash
      & \textbf{97.4\%}
      & 87.8\%
      & \textbf{85.5\%}
      & \(1.00\times\) \\
    Gemini 3.5 Flash
      & 92.7\%
      & \textbf{88.3\%}
      & 81.9\%
      & \(7.21\times\) \\
    GPT-5.4
      & 87.1\%
      & 84.0\%
      & 73.2\%
      & \(1.93\times\) \\
    Claude Sonnet 5
      & 78.9\%
      & 80.5\%
      & 63.5\%
      & \(3.05\times\) \\
    \bottomrule
  \end{tabular}
\end{table}

As shown in Table~\ref{tab:backbone}, Gemini 2.5 Flash provides the
strongest overall quality--coverage--cost trade-off. Although Gemini
3.5 Flash increases extraction coverage by \(0.5\) percentage points,
its lower judge-supported rate reduces publication coverage from
85.5\% to 81.9\%, while increasing inference cost by more than
\(7\times\). GPT-5.4 and Claude Sonnet 5 achieve still lower
publication coverage, at 73.2\% and 63.5\%, respectively.

Error analysis shows that the lower judge-supported rates of the
alternative backbones arise primarily from unsupported or partial
extractions rather than explicit contradictions or hallucinations.
These errors are concentrated in evidence-intensive attributes,
including unit count and free-text descriptions. The results
therefore show that backbone choice affects not only how often the
\texttt{ScoutAgent} extracts a value, but also how often that value is
sufficiently grounded for automatic publication.

Because all \texttt{ScoutAgent} variants are evaluated using the same fixed
\texttt{JudgeAgent}, these results provide a controlled comparison of
operational behavior rather than estimates of human-validated
accuracy.

\section{Deployment}
\label{sec:deployment}
We deployed TRACE in production and enriched 31 million SKUs across four business verticals. This increased impression-weighted enrichment
coverage, defined as the share of customer impressions associated with product
records carrying enriched attributes, by over 90\% across these
verticals.

\begin{figure}[th]
  \centering
  \includegraphics[width=\linewidth]{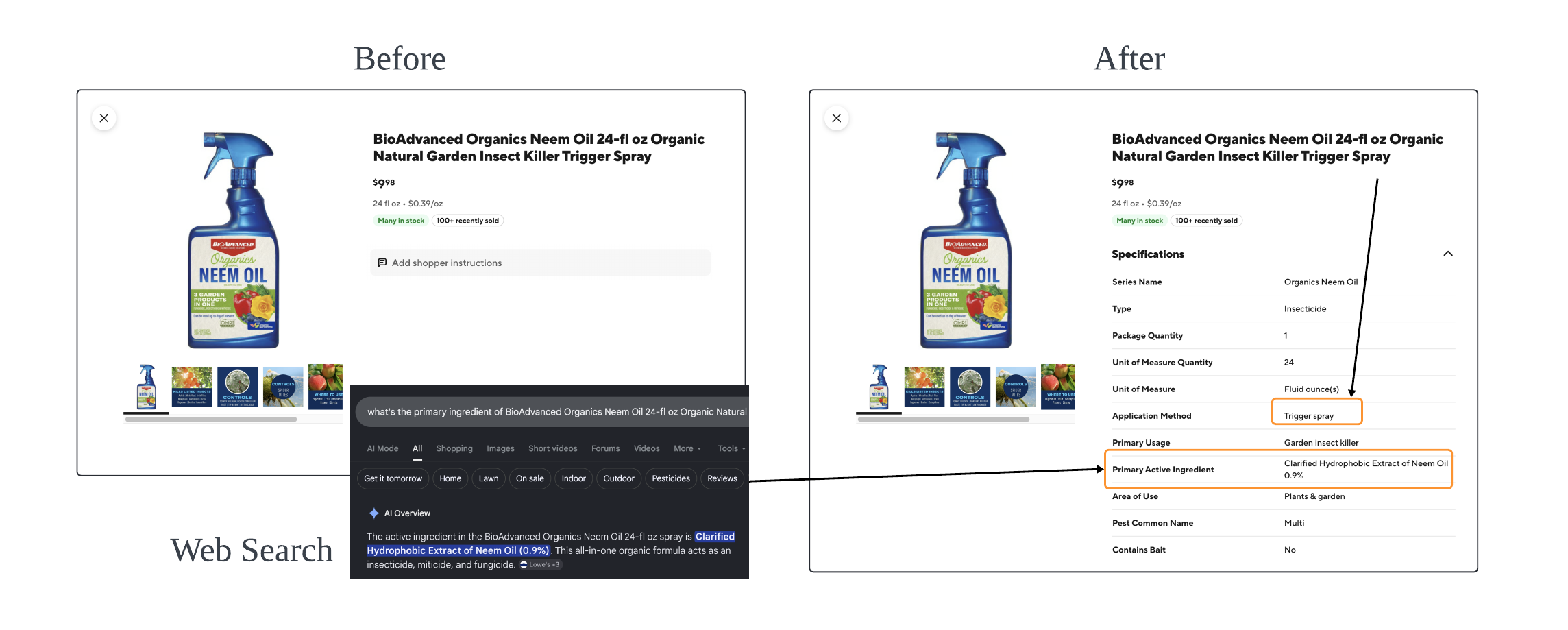}
  \caption{Illustrative product detail page (PDP) before and after catalog
  enrichment. Before enrichment, attributes were buried in unstructured text, making it difficult for shoppers to make a confident purchasing decision. TRACE generates structured attribute data using both internal and external data source.}
  \label{fig:pdp-example}
\end{figure}

\subsection{User Impact}
\label{sec:pdp}
\textbf{Product Detail Page} (PDP) presents the information shoppers use to
evaluate a product before adding it to their cart or completing a
purchase. When catalog attributes are missing, shoppers may have
difficulty determining whether a product satisfies their needs or
may form expectations that do not match the item ultimately
received. We hypothesize that surfacing enriched attributes on the
PDP reduces this information gap, leading to more confident purchase
decisions and fewer post-purchase issues.

We evaluate this hypothesis through a five-week randomized A/B test
with 90\% of traffic assigned to treatment and 10\% to holdout. In
the treatment, shoppers were shown PDPs augmented with the
attributes produced by TRACE; the holdout retained the existing PDP
experience. An example PDP before and after the experiment is shown in
\autoref{fig:pdp-example}. The experiment therefore measures the
end-to-end user impact of generating, validating, and surfacing
enriched catalog information.

\begin{table}[t]
  \caption{Online A/B test of surfacing enriched product detail pages
(PDPs). Enriched PDPs improved shopping outcomes by increasing checkout
conversion, with larger gains among power users, and reducing the rate
of missing or incorrect items. Effects are reported as relative changes versus the
control group, with 95\% confidence intervals and \(p\)-values.}
  \label{tab:ab}
  \centering
  \small
  \begin{tabular}{@{}lrrr@{}}
    \toprule
    Metric & Effect (rel.) & 95\% CI & \emph{p}-value \\
    \midrule
    Checkout conversion                       & +0.48\% & [+0.04\%, +0.92\%] & 0.034 \\
    Checkout conversion (power users)          & +1.18\% & [+0.24\%, +2.12\%] & 0.014 \\
    Missing/incorrect item rate  & −1.08\%          & [−2.04\%, −0.13\%] & 0.026 \\
    \bottomrule
  \end{tabular}
\end{table}

As shown in Table~\ref{tab:ab}, the enriched PDP increased
checkout conversion by \(0.48\%\), with a larger \(1.18\%\) increase
among power users. It also reduced the missing/incorrect-item rate by
\(1.08\%\). These results are consistent with the hypothesis that
richer product information helps shoppers make more informed
purchase decisions and form more accurate expectations about the
items they order.

\section{Limitations}
\label{sec:limits}
The Electronics and Home Improvement results use judge-supported
rate as a scalable operational metric rather than as a substitute for
human-validated precision. The JudgeAgent was calibrated on the
Grocery and Alcohol human audit, while its transfer to other
categories has received more limited human evaluation. Moreover,
because the ScoutAgent and JudgeAgent use models from the same family
in the primary configuration, they may exhibit correlated failure
modes.

The online experiment evaluates the end-to-end effect of displaying
enriched product pages. It therefore demonstrates the value of the
deployed system as a whole, but does not isolate the contribution of
the JudgeAgent or write-gating policy.

\section{Conclusion}
\label{sec:conclusion}

We presented TRACE, a multi-agent framework for evidence-grounded
catalog attribute enrichment. TRACE separates candidate generation
from verification: a \texttt{ScoutAgent} gathers and reconciles evidence from
multiple sources, while a \texttt{JudgeAgent} applies a stricter evidence
policy before proposed values are written to the catalog.

Experiments across multiple business verticals demonstrate that
TRACE produces high-quality attribute values and can operate at
production scale. On the human-annotated dataset, TRACE achieved
98.2\% accuracy. A randomized online experiment showed that
surfacing enriched attributes on product detail pages increased
checkout conversion by 0.48\% and reduced missing or incorrect item
reports.

These findings show that evidence-grounded catalog enrichment can
improve both catalog quality and downstream user experience. More
broadly, they highlight the importance of grounding generated
attribute values in product-specific evidence and verifying them
before publication.


\section*{Declaration on Generative AI}
During the preparation of this work, the author(s) used generative AI tools in
order to: Grammar and spelling check; and Paraphrase and reword. After using
these tool(s)/service(s), the author(s) reviewed and edited the content as
needed and take(s) full responsibility for the publication's content.

\bibliography{refs}

@inproceedings{zheng2023mtbench,
  title        = {Judging {LLM}-as-a-Judge with {MT}-Bench and Chatbot Arena},
  author       = {Zheng, Lianmin and Chiang, Wei-Lin and Sheng, Ying and Zhuang, Siyuan and Wu, Zhanghao and Zhuang, Yonghao and Lin, Zi and Li, Zhuohan and Li, Dacheng and Xing, Eric P. and Zhang, Hao and Gonzalez, Joseph E. and Stoica, Ion},
  booktitle    = {Advances in Neural Information Processing Systems 36 (NeurIPS 2023) Datasets and Benchmarks Track},
  year         = {2023},
  eprint       = {2306.05685},
  archivePrefix = {arXiv},
  primaryClass = {cs.CL},
  note         = {arXiv:2306.05685}
}

@inproceedings{liu2023geval,
  title        = {{G-Eval}: {NLG} Evaluation using {GPT-4} with Better Human Alignment},
  author       = {Liu, Yang and Iter, Dan and Xu, Yichong and Wang, Shuohang and Xu, Ruochen and Zhu, Chenguang},
  booktitle    = {Proceedings of the 2023 Conference on Empirical Methods in Natural Language Processing (EMNLP)},
  pages        = {2511--2522},
  year         = {2023},
  publisher    = {Association for Computational Linguistics},
  doi          = {10.18653/v1/2023.emnlp-main.153}
}

@inproceedings{saadfalcon2024ares,
  title        = {{ARES}: An Automated Evaluation Framework for Retrieval-Augmented Generation Systems},
  author       = {Saad-Falcon, Jon and Khattab, Omar and Potts, Christopher and Zaharia, Matei},
  booktitle    = {Proceedings of the 2024 Conference of the North American Chapter of the Association for Computational Linguistics: Human Language Technologies (NAACL-HLT)},
  year         = {2024},
  publisher    = {Association for Computational Linguistics},
  eprint       = {2311.09476},
  archivePrefix = {arXiv},
  primaryClass = {cs.CL},
  note         = {arXiv:2311.09476}
}

@inproceedings{es2024ragas,
  title        = {{RAGAs}: Automated Evaluation of Retrieval Augmented Generation},
  author       = {Es, Shahul and James, Jithin and Espinosa-Anke, Luis and Schockaert, Steven},
  booktitle    = {Proceedings of the 18th Conference of the European Chapter of the Association for Computational Linguistics (EACL): System Demonstrations},
  pages        = {150--158},
  year         = {2024},
  publisher    = {Association for Computational Linguistics},
  doi          = {10.18653/v1/2024.eacl-demo.16}
}

@misc{verga2024poll,
  title        = {Replacing Judges with Juries: Evaluating {LLM} Generations with a Panel of Diverse Models},
  author       = {Verga, Pat and Hofst{\"a}tter, Sebastian and Althammer, Sophia and Su, Yixuan and Piktus, Aleksandra and Arkhangorodsky, Arkady and Xu, Minjie and White, Naomi and Lewis, Patrick},
  year         = {2024},
  eprint       = {2404.18796},
  archivePrefix = {arXiv},
  primaryClass = {cs.CL},
  note         = {arXiv:2404.18796}
}

@inproceedings{zheng2018opentag,
  title        = {{OpenTag}: Open Attribute Value Extraction from Product Profiles},
  author       = {Zheng, Guineng and Mukherjee, Subhabrata and Dong, Xin Luna and Li, Feifei},
  booktitle    = {Proceedings of the 24th ACM SIGKDD International Conference on Knowledge Discovery \& Data Mining (KDD)},
  year         = {2018},
  publisher    = {ACM},
  doi          = {10.1145/3219819.3219839}
}

@inproceedings{dong2020autoknow,
  title        = {{AutoKnow}: Self-Driving Knowledge Collection for Products of Thousands of Types},
  author       = {Dong, Xin Luna and He, Xiang and Kan, Andrey and Li, Xian and Liang, Yan and Ma, Jun and Xu, Yifan Ethan and Zhang, Chenwei and Zhao, Tong and Blanco Saldana, Gabriel and Deshpande, Saurabh and Manduca, Alexandre Michetti and Ren, Jay and Singh, Surender Pal and Xiao, Fan and Chang, Haw-Shiuan and Karamanolakis, Giannis and Mao, Yuning and Wang, Yaqing and Faloutsos, Christos and McCallum, Andrew and Han, Jiawei},
  booktitle    = {Proceedings of the 26th ACM SIGKDD International Conference on Knowledge Discovery \& Data Mining (KDD)},
  year         = {2020},
  publisher    = {ACM},
  doi          = {10.1145/3394486.3403323},
  eprint       = {2006.13473},
  archivePrefix = {arXiv},
  primaryClass = {cs.AI}
}

@inproceedings{zhang2021queaco,
  title        = {{QUEACO}: Borrowing Treasures from Weakly-labeled Behavior Data for Query Attribute Value Extraction},
  author       = {Zhang, Danqing and Li, Zheng and Cao, Tianyu and Luo, Chen and Wu, Tony and Lu, Hanqing and Song, Yiwei and Yin, Bing and Zhao, Tuo and Yang, Qiang},
  booktitle    = {Proceedings of the 30th ACM International Conference on Information and Knowledge Management (CIKM)},
  year         = {2021},
  publisher    = {ACM},
  doi          = {10.1145/3459637.3481946},
  eprint       = {2108.08468},
  archivePrefix = {arXiv},
  primaryClass = {cs.CL}
}

@misc{sinha2024pae,
  title        = {{PAE}: {LLM}-based Product Attribute Extraction for E-Commerce Fashion Trends},
  author       = {Sinha, Apurva and Gujral, Ekta},
  year         = {2024},
  eprint       = {2405.17533},
  archivePrefix = {arXiv},
  primaryClass = {cs.CL},
  note         = {arXiv:2405.17533}
}

@misc{lu2025karma,
  title        = {{KARMA}: Leveraging Multi-Agent {LLMs} for Automated Knowledge Graph Enrichment},
  author       = {Lu, Yuxing and Wang, Jinzhuo},
  year         = {2025},
  eprint       = {2502.06472},
  archivePrefix = {arXiv},
  primaryClass = {cs.CL},
  note         = {arXiv:2502.06472}
}

@inproceedings{zhang2025patternrag,
  title        = {Leveraging Product Catalog Patterns for Multilingual E-commerce Product Attribute Prediction},
  author       = {Zhang, Bryan and Khan, Suleiman A. and Walter, Stephan},
  booktitle    = {Proceedings of the 2025 Conference on Empirical Methods in Natural Language Processing: Industry Track},
  pages        = {267--275},
  year         = {2025},
  publisher    = {Association for Computational Linguistics},
  doi          = {10.18653/v1/2025.emnlp-industry.18}
}

@misc{shi2025searchrag,
  title        = {{SearchRAG}: Can Search Engines Be Helpful for {LLM}-based Medical Question Answering?},
  author       = {Shi, Yucheng and Yang, Tianze and Chen, Canyu and Li, Quanzheng and Liu, Tianming and Li, Xiang and Liu, Ninghao},
  year         = {2025},
  eprint       = {2502.13233},
  archivePrefix = {arXiv},
  primaryClass = {cs.CL},
  note         = {arXiv:2502.13233}
}

@misc{khandelwal2023generative,
  title        = {Large Scale Generative Multimodal Attribute Extraction for E-commerce Attributes},
  author       = {Khandelwal, Anant and Mittal, Happy and Kulkarni, Shreyas Sunil and Gupta, Deepak Kumar},
  year         = {2023},
  eprint       = {2306.00379},
  archivePrefix = {arXiv},
  primaryClass = {cs.CL},
  note         = {arXiv:2306.00379}
}

@misc{huang2025madiave,
  title        = {{MADIAVE}: Multi-Agent Debate for Implicit Attribute Value Extraction},
  author       = {Huang, Wei-Chieh and Caragea, Cornelia},
  year         = {2025},
  eprint       = {2510.05611},
  archivePrefix = {arXiv},
  primaryClass = {cs.CL},
  note         = {arXiv:2510.05611}
}

@misc{wen2025knowlimitssurveyabstention,
      title={Know Your Limits: A Survey of Abstention in Large Language Models}, 
      author={Bingbing Wen and Jihan Yao and Shangbin Feng and Chenjun Xu and Yulia Tsvetkov and Bill Howe and Lucy Lu Wang},
      year={2025},
      eprint={2407.18418},
      archivePrefix={arXiv},
      primaryClass={cs.CL},
      url={https://arxiv.org/abs/2407.18418}, 
}

@inproceedings{yao2023react,
  title        = {{ReAct}: Synergizing Reasoning and Acting in Language Models},
  author       = {Yao, Shunyu and Zhao, Jeffrey and Yu, Dian and Du, Nan and Shafran, Izhak and Narasimhan, Karthik and Cao, Yuan},
  booktitle    = {The Eleventh International Conference on Learning Representations (ICLR)},
  year         = {2023},
  eprint       = {2210.03629},
  archivePrefix = {arXiv},
  primaryClass = {cs.CL},
  note         = {arXiv:2210.03629}
}

@inproceedings{nigam2019semantic,
  title        = {Semantic Product Search},
  author       = {Nigam, Priyanka and Song, Yiwei and Mohan, Vijai and Lakshman, Vihan and Ding, Weitian and Shingavi, Ankit and Teo, Choon Hui and Gu, Hao and Yin, Bing},
  booktitle    = {Proceedings of the 25th ACM SIGKDD International Conference on Knowledge Discovery \& Data Mining (KDD)},
  year         = {2019},
  publisher    = {ACM},
  doi          = {10.1145/3292500.3330759},
  eprint       = {1907.00937},
  archivePrefix = {arXiv},
  primaryClass = {cs.IR}
}

@inproceedings{wang2019kgat,
  title        = {{KGAT}: Knowledge Graph Attention Network for Recommendation},
  author       = {Wang, Xiang and He, Xiangnan and Cao, Yixin and Liu, Meng and Chua, Tat-Seng},
  booktitle    = {Proceedings of the 25th ACM SIGKDD International Conference on Knowledge Discovery \& Data Mining (KDD)},
  year         = {2019},
  publisher    = {ACM},
  doi          = {10.1145/3292500.3330989},
  eprint       = {1905.07854},
  archivePrefix = {arXiv},
  primaryClass = {cs.IR}
}

@inproceedings{yang2022mave,
  title        = {{MAVE}: A Product Dataset for Multi-source Attribute Value Extraction},
  author       = {Yang, Li and Wang, Qifan and Yu, Zac and Kulkarni, Anand and Sanghai, Sumit and Shu, Bin and Elsas, Jon and Kanagal, Bhargav},
  booktitle    = {Proceedings of the Fifteenth ACM International Conference on Web Search and Data Mining (WSDM)},
  pages        = {1256--1265},
  year         = {2022},
  publisher    = {ACM},
  doi          = {10.1145/3488560.3498377},
  eprint       = {2112.08663},
  archivePrefix = {arXiv},
  primaryClass = {cs.CL}
}

@inproceedings{zhu2020mjave,
  title        = {Multimodal Joint Attribute Prediction and Value Extraction for {E}-commerce Product},
  author       = {Zhu, Tiangang and Wang, Yue and Li, Haoran and Wu, Youzheng and He, Xiaodong and Zhou, Bowen},
  booktitle    = {Proceedings of the 2020 Conference on Empirical Methods in Natural Language Processing (EMNLP)},
  pages        = {2129--2139},
  year         = {2020},
  publisher    = {Association for Computational Linguistics},
  doi          = {10.18653/v1/2020.emnlp-main.166}
}

@misc{brinkmann2023llmpave,
  title        = {ExtractGPT: Exploring the Potential of Large Language Models for Product Attribute Value Extraction},
  author       = {Brinkmann, Alexander and Shraga, Roee and Bizer, Christian},
  year         = {2023},
  eprint       = {2310.12537},
  archivePrefix = {arXiv},
  primaryClass = {cs.CL},
  note         = {arXiv:2310.12537}
}

@misc{comanici2025gemini25pushingfrontier,
      title={Gemini 2.5: Pushing the Frontier with Advanced Reasoning, Multimodality, Long Context, and Next Generation Agentic Capabilities}, 
      author={Gheorghe Comanici and Eric Bieber and Mike Schaekermann and Ice Pasupat and Noveen Sachdeva and Inderjit Dhillon and Marcel Blistein and Ori Ram and Dan Zhang and Evan Rosen and Luke Marris and Sam Petulla and Colin Gaffney and Asaf Aharoni and Nathan Lintz and Tiago Cardal Pais and Henrik Jacobsson and Idan Szpektor and Nan-Jiang Jiang and ... and Wesley Helmholz},
      year={2025},
      eprint={2507.06261},
      archivePrefix={arXiv},
      primaryClass={cs.CL},
      url={https://arxiv.org/abs/2507.06261}, 
}

\appendix

\section{Condensed Agent Prompt Templates}
\label{app:prompts}

The following provider-neutral templates reproduce the instruction contracts
used in the final configuration. They are condensed rather than verbatim:
repeated prose, concrete ontology values, SKU content, and provider-specific
tool schemas are omitted for space. Angle-bracketed fields are populated for
each SKU. TRACE makes one \texttt{ScoutAgent} call and one
\texttt{JudgeAgent} call per eligible SKU; each call returns a map of
per-attribute outputs.

\begin{figure}[p]
  \centering
  \begin{promptbox}{\texttt{ScoutAgent}: Condensed Instruction Contract}

You enrich ONE retail SKU. Extract every requested priority attribute using
available evidence; abstain rather than infer.

\medskip
\textbf{Evidence priority:} (1) catalog, (2) high-quality syndicated data,
(3) product image, and (4) web. Higher-priority evidence overrides
lower-priority evidence in a conflict; web fills gaps rather than correcting
catalog data.

\medskip
\textbf{Search:} \traceph{SEARCH\_TOOL} is expected for unresolved attributes.
Search iteratively in this order: UPC/GTIN/EAN; exact brand + product name +
size/variant; model/MPN; official manufacturer page; and then
attribute-specific queries. Prefer official product pages and manuals,
followed by major retailers, other retailers, and aggregators. Avoid blogs,
forums, reviews, and user-generated content.

\medskip
\textbf{Identity:} Before using image or web evidence, match the same brand,
product variant, size, and pack count. Product identifiers and model/MPN may
be used to establish the match. Do not use evidence from a nearby variant. If
an otherwise relevant source fails this check, return
\texttt{identity\_mismatch} for attributes that would rely on it.

\medskip
\textbf{Value constraints:} ENUM values must come from the supplied set;
STRING values should use the preferred vocabulary when supported; BOOLEAN
values are \texttt{"True"} or \texttt{"False"}. A multi-source value is
allowed only when each named source contributes distinct content.

\medskip
\textbf{Grounding:} Each extracted value includes
\texttt{value}, \texttt{source}, \texttt{source\_type},
\texttt{source\_ref}, \texttt{quoted\_evidence}, and
\texttt{confidence} in $[0,1]$. Evidence must be a verbatim span; at most
three short spans may be joined for a synthesized field.

\medskip
\textbf{Output:} Return valid JSON only. Every requested slug must appear:

\medskip
\texttt{\{"identity":\{"matched\_brand":...,}

\texttt{"matched\_variant":...,"matched\_size":...,}

\texttt{"matched\_pack":...,"confidence":0.0\},}

\texttt{"attributes":\{"\traceph{SLUG}":\{}

\texttt{"status":"extracted|not\_found|not\_applicable|}

\texttt{ambiguous|conflict|identity\_mismatch",}

\texttt{"value":...|null,}

\texttt{"source":"cd|hq|image|web"|[...],}

\texttt{"source\_type":"catalog|hq|image|official\_web|}

\texttt{official\_doc|major\_retailer|retailer|aggregator|other",}

\texttt{"source\_ref":...,}

\texttt{"quoted\_evidence":...,}

\texttt{"confidence":0.0\}\}\}}

\medskip
For a non-extracted status, set \texttt{value=null} and omit provenance and
confidence fields.

\medskip
\textbf{Per-SKU payload:}
\traceph{PRIORITY\_ATTRIBUTES} (slug, definition, example, data type, enum or
preferred values, and multi-value flag);
\traceph{SKU\_CONTEXT} (name, details, size, identifiers, brand, and
category); \traceph{CATALOG\_EVIDENCE};
\traceph{SYNDICATED\_EVIDENCE}; and \traceph{IMAGE}.
Extract all listed slugs for this SKU.

  \end{promptbox}

  \caption{Condensed \texttt{ScoutAgent} prompt template for the final TRACE
  configuration. Provider-specific message wrappers, tool schemas, concrete
  ontology values, and SKU data are represented by placeholders.}
  \label{fig:scout-prompt}
\end{figure}

\begin{figure}[p]
  \centering
  \begin{promptbox}{\texttt{JudgeAgent}: Condensed Instruction Contract}

You are a precise quality auditor with web access. For every
\texttt{(slug,value,status)} produced for ONE SKU, return one verdict:
\texttt{PASS}, \texttt{FAIL}, \texttt{UNVERIFIED}, or
\texttt{UNCERTAIN}.

\medskip
\textbf{Decision order:}

\medskip
1. For \texttt{not\_found}, \texttt{not\_applicable},
\texttt{ambiguous}, \texttt{conflict}, or
\texttt{identity\_mismatch}, PASS a reasonable abstention; FAIL with
\texttt{MISSING\_EXTRACTION} when the value is clearly extractable.

\medskip
2. If catalog, syndicated, or image evidence directly supports the value,
PASS without search. If it directly contradicts the value, FAIL with
\texttt{CONTRADICTION} without search.

\medskip
3. If the candidate is web-sourced and the supplied evidence is silent,
search \traceph{SEARCH\_TOOL} using brand + model/MPN + product name +
attribute, adding UPC/GTIN when available. Verify the exact brand, model,
variant, size/capacity, and pack. Exact-product confirmation is PASS; an
exact-product contradiction is FAIL/\texttt{INCORRECT\_VALUE}; a similar
variant, generic page, or no relevant result is \texttt{UNVERIFIED}.

\medskip
4. If a catalog-, syndicator-, or image-sourced citation does not support
the claim, return FAIL/\texttt{UNSUPPORTED\_INFERENCE}; do not rescue it
with web evidence.

\medskip
5. If exact-product sources actively conflict, return
\texttt{UNCERTAIN}.

\medskip
\textbf{FAIL versus UNVERIFIED:} FAIL requires positive evidence that the
claim is wrong or fabricated. Use UNVERIFIED when the claim is plausible
and uncontradicted but cannot be confirmed for the exact variant.

\medskip
\textbf{Normalization:} Treat Boolean synonyms, standard technical
abbreviations, and equivalent unit formats as equal. Accept minor numeric
rounding within 2\% or 0.5 absolute.

\medskip
\textbf{Free text:} Catalog and syndicator quotes must be verbatim. Web
claims must be factually supported. If a partial quote establishes the core
claim but not a benign continuation, return UNVERIFIED. If the unsupported
tail introduces a distinct factual claim, return
FAIL/\texttt{PARTIAL\_EXTRACTION}.

\medskip
\textbf{FAIL taxonomy:}
\texttt{INCORRECT\_VALUE}, \texttt{UNSUPPORTED\_INFERENCE},
\texttt{HALLUCINATION}, \texttt{CONTRADICTION},
\texttt{SCHEMA\_MALFORMED}, \texttt{PARTIAL\_EXTRACTION},
\texttt{INVALID\_FORMAT}, \texttt{UNSAFE}, or
\texttt{MISSING\_EXTRACTION}.

\medskip
\textbf{Output:} Return valid JSON only:

\medskip
\texttt{\{"verdicts":\{"\traceph{SLUG}":\{}

\texttt{"verdict":"PASS|FAIL|UNVERIFIED|UNCERTAIN",}

\texttt{"failure\_type":null|"\traceph{FAILURE\_TYPE}",}

\texttt{"reasoning":"\traceph{AT MOST 30 WORDS}",}

\texttt{"url":null|"\traceph{EVIDENCE\_URL}"\}\}\}}

\medskip
\textbf{Per-SKU payload:} \traceph{SKU\_CONTEXT};
\traceph{CATALOG\_EVIDENCE}; \traceph{SYNDICATED\_EVIDENCE};
\traceph{IMAGE}; and \traceph{CANDIDATE\_VALUES}, containing the
\texttt{ScoutAgent}'s complete per-attribute JSON map. Judge every entry and
return the schema above.

  \end{promptbox}

  \caption{Condensed \texttt{JudgeAgent} prompt template for the final TRACE
  configuration. Provider-specific message wrappers, tool schemas, concrete
  ontology values, and SKU data are represented by placeholders.}
  \label{fig:judge-prompt}
\end{figure}

\end{document}